\documentclass[twocolumn]{revtex4}
\usepackage{graphicx}
\usepackage{amsmath} 
\usepackage{subcaption} 

\date{\today}

\def\usepdf{0}

\begin{document}
\title{Efficient forward propagation of time-sequences in convolutional neural networks using Deep Shifting}
\author{Koen Groenland, Sander Bohte}
\affiliation{Centrum Wiskunde en Informatica}
\email{groenlan@cwi.nl}

\begin{abstract} 
When a Convolutional Neural Network is used for on-the-fly evaluation of continuously updating time-sequences, many redundant convolution operations are performed. We propose the method of Deep Shifting, which remembers previously calculated results of convolution operations in order to minimize the number of calculations. The reduction in complexity is at least a constant and in the best case quadratic. We demonstrate that this method does indeed save significant computation time in a practical implementation, especially when the networks receives a large number of time-frames.
\end{abstract}

\maketitle

\section{Introduction}
Convolutional Neural Networks (CNNs) have proven extremely succesful in finding structure in high-dimensional data, including time-sequences such as audio and video \cite{Baccouche2011, Ji2010, Sainath2013}. Some examples of promising real-world applications are speech- and video recognition, and automatic translation \cite{LeCun2015}. Challenges for software development include training of the networks using large GPU clusters and gathering huge, labeled datasets \cite{Bengio2012}. Practical user-side evaluation faces completely different challenges, including efficient and fast performance, low resource consumption, and responsiveness, such that the software responds to recognized events as quickly as possible \cite{Han2015}. Earlier work focusing on achieving these challenges include using less-parameter convolution filters \cite{Szegedy2015}, pruning obsolete weights \cite{Han2015}, and using spiking networks \cite{Bohte2012}. This paper deals with optimizing convolution of time-series, as used for example in 3D convolutional neural networks as they are applied in human action recognition \cite{Ji2010}. 

We observe that when forward propagating continuously updating time-sequences through a neural network that applies convolution in the time-dimension, many redundant calculations are made. In order to avoid these calculations, to save CPU resources and potentially battery life on mobile devices, we propose \textit{Deep Shifting}, which copies results of convolution operations from earlier time steps, rather than re-calculating these over and over. This can save substantial calculation time, especially when the CNN looks at a large number of time-frames. 
This paper is organized as follows: Section 2 shows how Deep Shifting performs CNN operations on time-sequences without performing redundant calculations. Sections 3 and 4 examine the theoretical and practical benefits of Deep Shifting. Section 5 investigates the possibilities of training a network using a minimal number of neurons and operations, and the paper finishes with a discussion and conclusion.

\section{Deep Shifting}
\begin{figure}[!b]
	\centering
	\includegraphics[width=0.4\columnwidth]{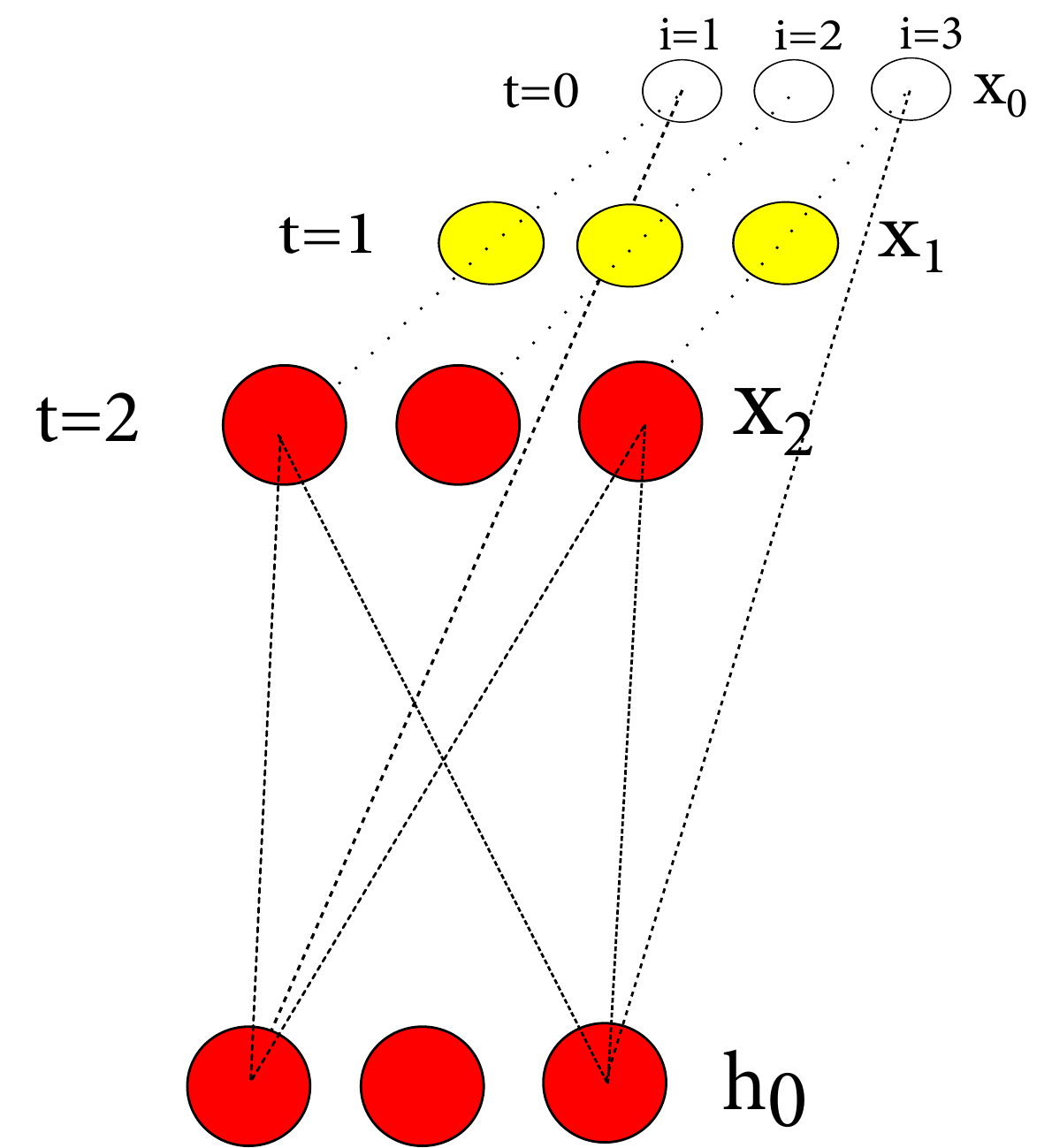}
	\caption{One timestep within the hidden layer is calculated by taking the weighted sum over all neurons within the convolution window. }
	\label{fig:conv_small}
\end{figure}

Let $x$ be an input with a matrix shape, having a time-axis and a context axis. The context axis holds what is called ``channels'' in image convolutions. A convolutional auto-encoder, for which we label the layers $x$, $h$ and $y$ respectively, applies the following encoding operation with convolution over the time axis:
\begin{equation}
h_t = \sigma \left( \sum_{\tau=0}^w W_\tau \cdot x_{t+\tau} + b \right).
\label{time_encoding}
\end{equation}
Here, $w$ is the size of the convolution window, $W$ is a set of $w$ weight matrices, $t$ denotes the time step of the input sequence, and $\tau$ labels the time axis of the weights. $\sigma$ denotes an activation function, which we choose to be $\tanh$ in our computational part. A schematic view is given in figure \ref{fig:conv_small}. 
\begin{figure}
  \centering
  \includegraphics[width=0.4\columnwidth]{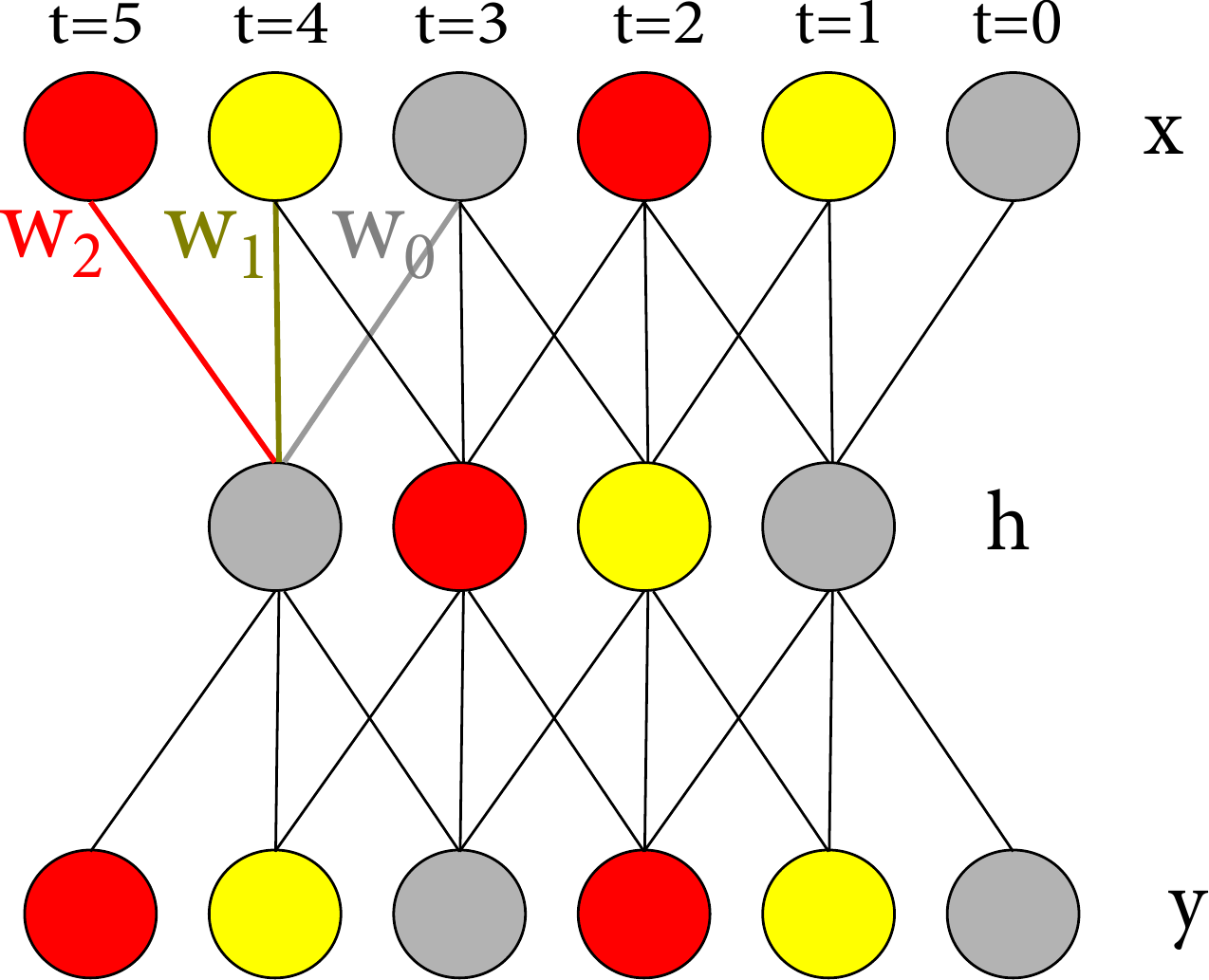}
  \caption{A graphical view of an auto-encoder which applies convolution on the time-axis of its input. }
  \label{fig:conv_normal}
\end{figure}

\begin{figure}
  \centering
  \includegraphics[width=0.6\columnwidth]{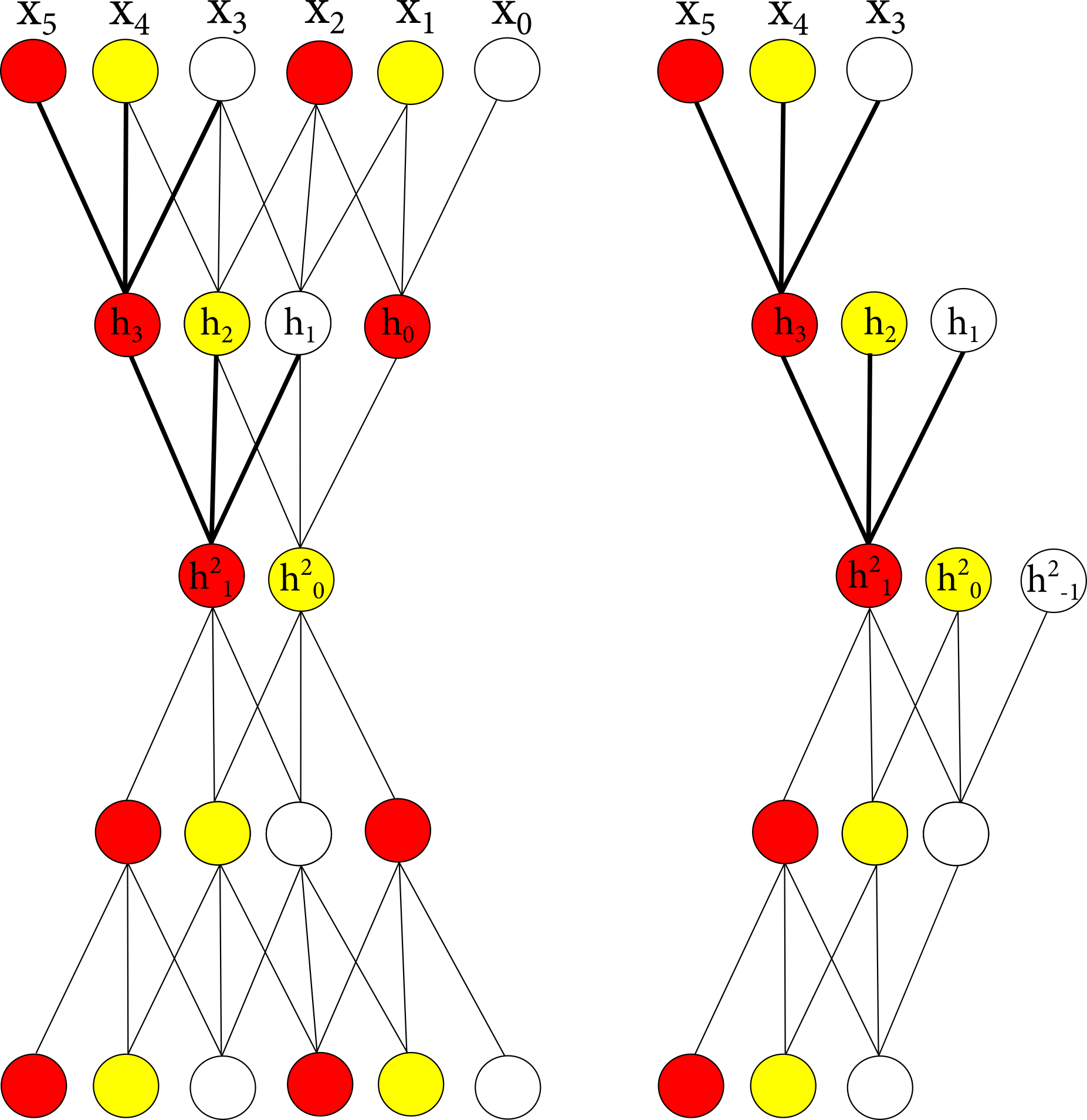}
  \caption{A comparison between a regular convolution network (left) and the pruned version version using Deep Shifting (right).  }
  \label{fig:comparison}
\end{figure}

If the input consists of time-evolving data, we typically want to perform the encoding every time new information becomes available, e.g. at every new time step. However, due to the nature of convolution, many redundant calculations are made. Let the actual time at which we perform an encoding be denoted by $T$, and denote the neuronal activation at a specific time step by $x_t(T)$ and $h_t(T)$. Moreover, let the indices $t$ be defined relative to $T$, such that the `leftmost' input neuron is $x_T$. If at time $T = 5$, the network in figure \ref{fig:conv_normal} received input time steps 0 through 5 initially, it has calculated $h_0(5)$ up to $h_3(5)$ through formula \ref{time_encoding}. When we give the same network time steps 1 through 6 at a later time $T=6$, it will calculate $h_1(6)$ through $h_4(6)$. By formula \ref{time_encoding}, $h_1(5) = h_1(6)$, and the same goes for the other time steps $h_2$ and $h_3$. By the natural flow of time, the activations in the network will have \textit{shifted} through the network. Therefore, we could have just as well stored these values, and copied them to the neighbouring neurons, without performing the the convolution calculation all over again. Let $t_h$ denote the highest time step a hidden neuron can have at the time of encoding, e.g. $t_h = T-w+1$. Using the Deep Shifting method, the neural activations are calculated by: 
\begin{eqnarray}
h_{t_h} &=& \sigma \left( \sum_{\tau=0}^{w} W_\tau \cdot x_{t_h+\tau} + b \right), \label{shift_encode} \\
h_t(T) &=& h_{t+1}(T-1) \hspace{1.5cm} (t < t_h). \label{shift_shift}
\end{eqnarray}
The same reasoning holds for an arbitrary number of layers, can be used to save calculations in down-sampling, and is easily extended to convolution in multiple dimensions, such as 3D convolutions that are often used on video data \cite{Ji2010}. Moreover, it can avoid obsolete reconstruction operations: Layer $y$ in figure \ref{fig:conv_normal} will have $y_3$ fixed after $T=5$, such that this value could also in principle be stored and shifted (assuming the weights will not have changed significantly in the meantime). A graphical comparison is given in figure \ref{fig:comparison}.

\section{Theoretical Benefit}
How much exactly did we prune from the original network? Let one ``convolution operation'' be the calculation of one time-frame in the hidden layer, e.g. one full vector $h_t$ for some $t$. If the layers $x$, $h$, $y$ would normally span $t_x$, $t_h$ and $t_y$ time steps, we don't have to store $t_x - w$, $t_h - w$ and $t_y - w$ neurons in the three layers, and we saved ourselves calculating $t_h - 1$ hidden convolution operations and $t_y - w$ output convolutions. 

Recall that for most practical applications, convolutional auto-encoders are stacked in an hourglass shape, where the shallow layers span many time steps, whilst the deeper layers have increasingly smaller time axes. Assume we start with the deepest hidden layer of a CAE, with a time-axis of size $t_{h^n}$, on which we will stack some number $n$ of shallower layers. Any previous layers should then have a time-axis of size $t_h + w - 1$ if a convolution window of size $w$ is used. The number of convolution operations $C$ performed in a forward pass through this network can be calculated to be:

\begin{align}
C_{\textnormal{normal}} = \ \ &t_{h^n} \nonumber \\
	&+ t_{h^n} + w - 1  \nonumber \\
	&+ t_{h^n} + w - 1 + w - 1  \nonumber  \\
	&+ t_{h^n} + w - 1 + w - 1 + w - 1 \nonumber  \\
	&+ \dots \nonumber  \\
	&+ t_{h^n} + (n-1)(w-1) \nonumber  \\
	=\ \ & n t_{h^n} + \frac{1}{2} (n-1)(n-2) (w-1).
\end{align}
In the last line, we used $\sum_{i=1}^n i = \frac{1}{2} n (n-1)$, where in our case the sum runs up to $(n-1)$ because the shallowest layer's neurons are not calculated through convolution. Clearly, due to the increasing size of the time-axes, the number of computations in a regular time-encoding CNN scales \textit{quadratically} with the number of layers, assuming fixed $t_{h^n}$. This case assumes the same $w$ in every layer, but this quadratic scaling holds even with $w$ varying per layer, as long as the windows have sizes of at least $2$. On the other hand, a Deep Shifting architecture scales linearly with the number of layers:
\begin{align}
C_{\textnormal{deep shifting}} = \ \ n.
\end{align}

The situation is slightly different if we consider the case where the time-axis of the input $t_x$ is fixed, and we add increasing numbers of \textit{deeper layers}, which have smaller time axes. In this case, the total number of convolutions upon encoding equals 
\begin{align}
C_{\textnormal{normal}} = \ \ &t_x - w + 1 \nonumber \\
	&+ t_x - w + 1 - w + 1 \nonumber \\
	&+ t_x - w + 1 - w + 1 - w + 1 \nonumber \\ 
	&+ \dots \nonumber \\ 
	&+ t_x - n (w-1) \nonumber \\
	= \ \ &n t_x - \frac{1}{2} n(n-1) (w-1).
\end{align}
Under these assumptions, Deep Shifting uses a number of convolution operations still equaling the number of layers. In this situation, we obtain a linear gain in performance: the average number of convolutions per layer is $t_x - \frac{n+1}{2} (w-1)$, compared to just $1$ for Deep Shifting. Therefore, Deep Shifting requires $t_x - \frac{n+1}{2} (w-1)$ times less convolution operations in the whole encoding process.

\section{Empirical Benefit}
\begin{figure}
  \centering
  \includegraphics[width=1.1\columnwidth]{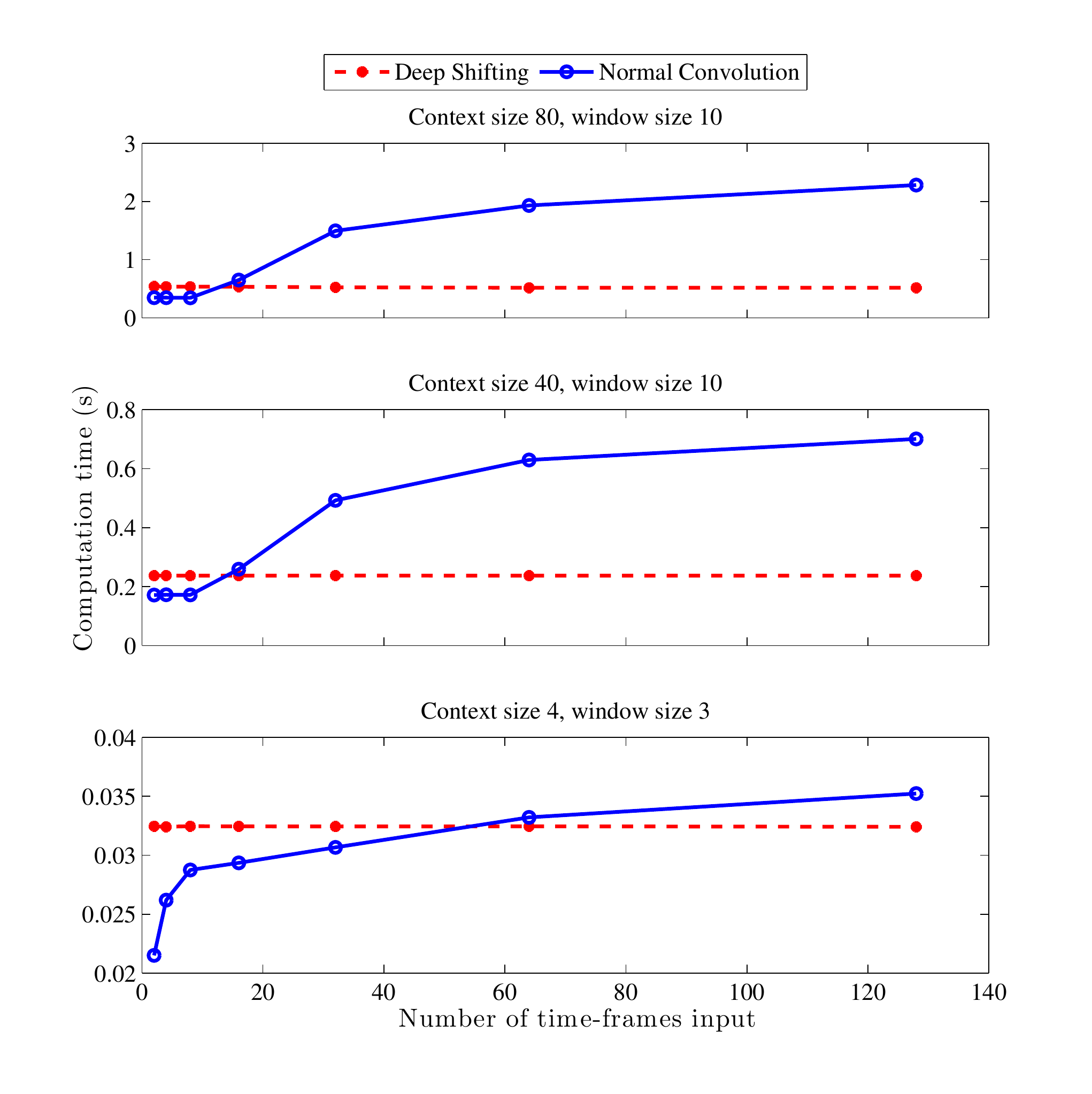}
  \caption{A comparison of computation time between ordinary CNNs, and CNNs that exploit Deep Shifting. All networks consisted of two convolution layers, each with the same number of output mappings. The results are the averages of 10 runs. }
  \label{fig:results}
\end{figure}

We compare our own Matlab implementation of a Deep Shifting network with the CNN from the `Deep Learn Toolbox' by Rasmus Bergpalm \cite{Bergpalm}. As the original version applies convolution in 2 dimensions for images specifically, we modified the code to apply convolution in only 1 dimension, representing time. We consider two networks: the first using the conventional CNN architecture (formula \ref{time_encoding}), the second exploiting Deep Shifting (formula \ref{shift_encode} and \ref{shift_shift}). Both networks receive a dataset representing part of a long time-sequence, and each subsequent input is the previous sequence shifted by one `neuron' in the time dimension. The average time required for forward propagation using various context axis sizes and convolution windows are displayed in figure \ref{fig:results}.
 
As expected, the computation time on the regular CNN increases with increasing time-window input $t_x$, whereas it remains constant using Deep Shifting. The shifting operation as implemented however requires a higher base computation time, and whether Deep Shifting does save computation time is then dependent on the parameters of the CNN. For realistic situations, such as a context size of 40 such as used in large-scale speech tasks \cite{Sainath2013}, Deep Shifting becomes beneficial when more than roughly 20 time-frames are given as input to the CNN. Moreover, the advantage of Deep Shifting becomes more noticeable when the size of the context axis increases.


\section{Training with minimal connectivity}
Can we also use the sparse connectivity structure sufficient for Deep Shifting in the context of learning?
This section compares the \textit{training} behaviour of Deep Shifting and a regular convolutional network. We train the networks as a Convolutional Auto-Encoder (CAE). Given an input $x$, the hidden layer $h_t$ is calculated using formula \ref{time_encoding} (normal CAE) or formulas \ref{shift_encode} and \ref{shift_shift} (Deep Shifting), using `valid convolution' (the number of time steps in the hidden layer is $t_h = t_x - w + 1$). Next, the network performs a reconstruction step, applying the adjoint of the earlier convolution to find the reconstruction $\tilde{x}$. This involves `full convolution', where the number of output time steps equals $t_{\tilde{x}} = t_h + w - 1 = t_x$. The network weight and bias parameters are set to minimize the reconstruction error $E = \sum_{x \in \textnormal{Dataset}} | x - \tilde{x} | ^2$ by using gradient descent for 100 epochs at a learning rate of $10^{-4}$. 

Our test involves 2 datesets. The first is the Spoken Digit dataset, as used in Ref. \cite{Verstraeten2005}, consisting of the number `zero' to `nine' pronounced 10 times by 5 different speakers. The second dataset is the Auslan dataset, a time-series representing hand-movements of the Australian deaf-community sign language, obtained from Ref. \cite{Lichman2013}. We test whether our networks can classify the right digit or sign, respectively. First, an MLP with 30 hidden units is trained to properly classify these on the \textit{training set}, after which the percentage of correct classifications on a separate \textit{test dataset} is measured. 

\begin{figure*}
\centering
\ifnum\usepdf=0  \includegraphics[height=9cm]{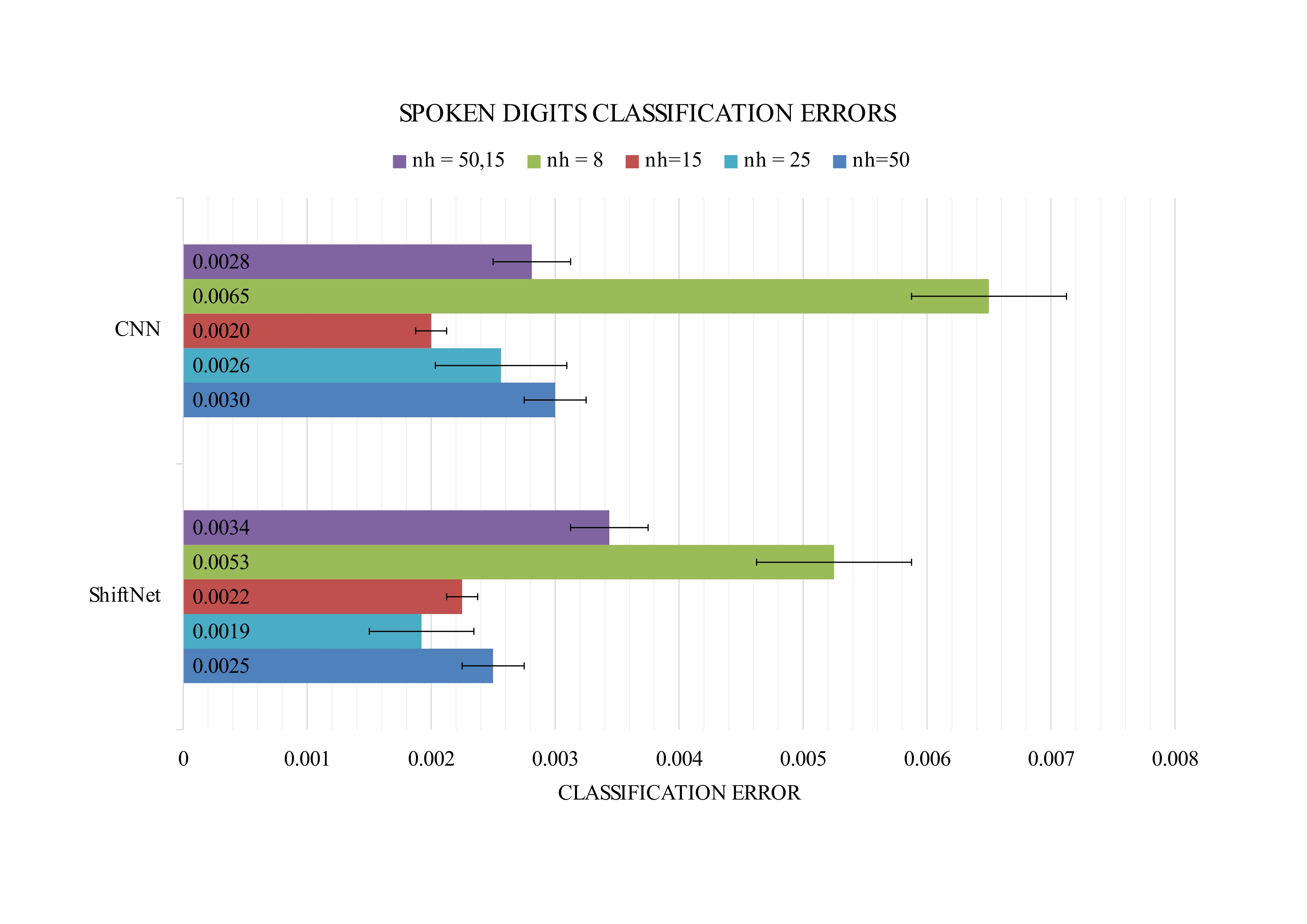} \fi
\caption{Results of the Spoken Digits classification test, showing the classification error (as a fraction of the total number of classifications) of a Deep Shifting network and a regular CNN with various hidden layer sizes. The purple bar indicated with `nh = 50,15' used two layers with 50 and 15 hidden neurons respectively. Displayed results are the averages of at least 16 runs, and the error bars indicate the standard error of the mean.}
\label{fig:class_vowel}
\end{figure*}

Figure \ref{fig:class_vowel} shows the results on the Spoken Digit dataset. The network denoted by `ShiftNet', which employed Deep Shifting, shows very competitive results compared to a regular CNN. Note that both networks manage to achieve a very low percentage of erroneous classifications, indicating that the networks encode all important information in the dataset into a small number of hidden units. Similarly to the approach of Ref. \cite{Verstraeten2005}, we used a random 60\% of the 500 data points for training, and the others for testing.

\begin{figure*}
\begin{subfigure}{\textwidth}
  \centering
\ifnum\usepdf=0  \includegraphics[height=9cm]{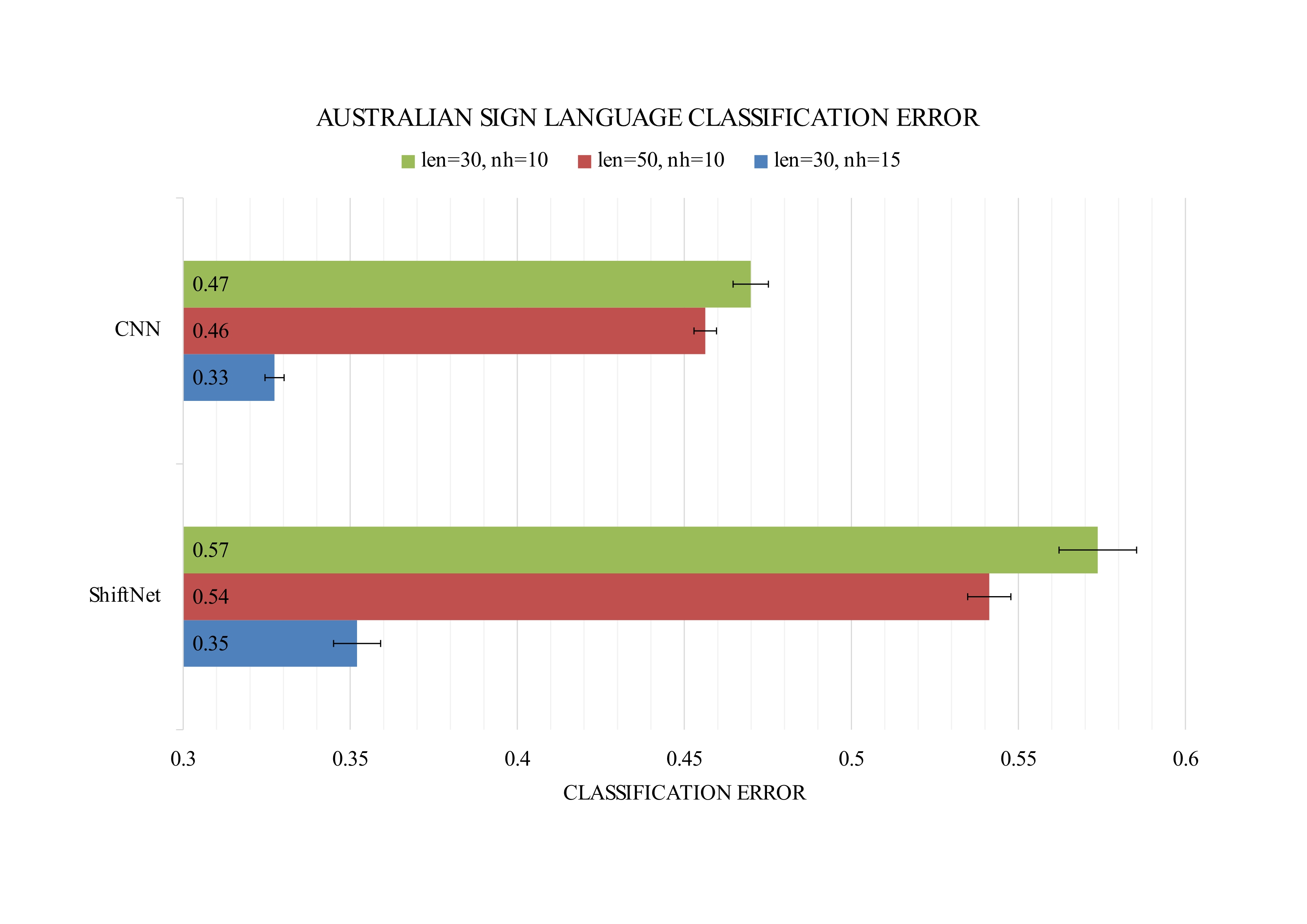} \fi
\end{subfigure}%
\caption{The classification errors on the Australian Sign Language dataset. The results are averages over multiple runs, and the networks were equipped with convolution windows of size 6. Error bars indicate the standard error of the mean. }
\label{fig:class_auslan}
\end{figure*}

Figure \ref{fig:class_auslan} shows the results on the Auslan dataset. The parameter `len' indicates the number of time-frames to which the input was scaled, and `nh' indicates the number of hidden units. On this much harder dataset, a regular CNN is able to perform much better than the ShiftNet, although the latter is still able to achieve very reasonable results despite the very limited connectivity. For this test, 10-fold cross validation is used, where 10\% of the data is taken as test data, and another 10\% used as validation data to limit overfitting by the MLP. 

In general, we observe that Deep Shifting is able to achieve results comparable to normal CNNs when the tests are `easy', for example, when the classification errors are low, or when many hidden units are used. When classification errors get large, or the number of hidden units is small compared to the data size, Deep Shifting generally does not reach the performance of a regular CNN.

\section{Discussion and Conclusion}
We described the method of Deep Shifting, which can speed up the forward propagation of continuously updating inputs in convolutional auto-encoders. Using both a theoretical and empirical analysis, we showed that Deep Shifting requires less convolution operations and computation time than a regular convolutional network when the number of input time-frames exceeds some threshold. For common practical applications, we found this threshold to lie around roughly 16 time-frames for 40 to 80 dimensional inputs with windows of size 10. Deep Shifting is only relevant when time-sequences need to be continuously evaluated, and when the number of time-frames to be considered is larger than the size of the convolution window. The latter will always be the case if multiple layers are used. 

Our analysis did not consider the use of graphical processing units: We are not sure how the speed of the copy operations of Deep Shifting compare against the large-scale parallel operations that GPU's are capable of, although we wonder if these can be used in practical user-end implementations. 

We also considered training a network with the Deep Shifting architecture. Our tests show that a regular neural network layout is preferred on hard training tasks. In general, training a Deep Shifting architecture has very few applications, since networks are often trained on readily stored datasets, limiting the need for on-the-fly evaluation. 

Still, Deep Shifting is an easily implementable trick to optimize deep neural networks working with time-evolving data, when speed or power consumption is relevant. This may be of benefit for mobile devices that aim to interpret sound, video, or other sensor data.

\end{document}